\documentclass[sigconf]{acmart}

\def\BibTeX{{\rm B\kern-.05em{\sc i\kern-.025em b}\kern-.08emT\kern-.1667em\lower.7ex\hbox{E}\kern-.125emX}}
    
\usepackage{tikz}
\usepackage{pgfplots}
\usepackage{float}
\usepackage{booktabs}

\copyrightyear{2020}
\acmYear{2020}
\setcopyright{acmcopyright}
\acmConference[CoDS COMAD 2020]{7th ACM IKDD CoDS and 25th COMAD}{January 5--7, 2020}{Hyderabad, India}
\acmPrice{15.00}
\acmDOI{10.1145/3371158.3371200}
\acmISBN{978-1-4503-7738-6/20/01}

\begin{document}

%
\title{On-Device Information Extraction from Screenshots in form of tags}

%
\author{Sumit Kumar\qquad Gopi Ramena\qquad Manoj Goyal\qquad Debi Mohanty\qquad Ankur Agarwal\qquad Benu Changmai\qquad Sukumar Moharana}
\email{sumit.kr@samsung.com, gopi.ramena@samsung.com, manoj.goyal@samsung.com, debi.m@samsung.com, ankur.a@samsung.com, b.changmai@samsung.com, msukumar@samsung.com}

\affiliation{Samsung Research Institute, Bangalore}

%
\renewcommand{\shortauthors}{Sumit and Gopi, et al.}

%
\begin{abstract}
We propose a method to make mobile screenshots easily searchable. In this paper, we present the workflow in which we: 1) pre-processed a collection of screenshots, 2) identified script present in image, 3) extracted unstructured text from images, 4) identified language of the extracted text, 5) extracted keywords from the text, 6) identified tags based on image features, 7) expanded tag set by identifying related keywords, 8) inserted image tags with relevant images after ranking and indexed them to make it searchable on device. We made the pipeline which supports multiple languages and executed it on-device, which addressed privacy concerns. We developed novel architectures for components in the pipeline, optimized performance and memory for on-device computation. We observed from experimentation that the solution developed can reduce overall user effort and improve end user experience while searching, whose results are published.
\end{abstract}

%
%


%

%

%
\maketitle

\section{Introduction}
"Screenshot/Screen capture" is the term often used to describe the action of capturing mobile screen to static image file. The extraction, representation and retrieval of screenshots in an effective manner will pave way for its use case in a number of cases, including social media, models of task switching, assessments of the role of fake news in democratic settings, etc. \cite{ReevesScreenomics}.
\begin{figure*}
  \centering
  \includegraphics[width=\linewidth]{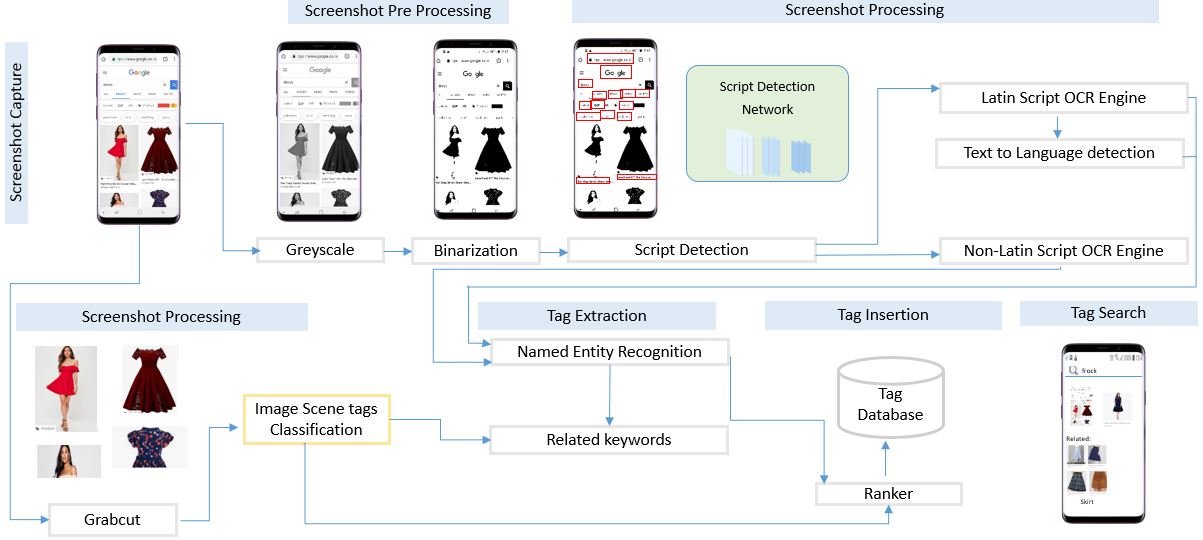}
  \caption{Overview of the Process Pipeline}
  \label{fig:pipeline}
\end{figure*}

As discussed in \cite{ChiattiTextExt} , screenshots provide a unique combination of graphic and scene text, motivating the evaluation of state-of-the-art OCR methods. Further, information extraction from digital screenshots raises the need for developing a general-purpose framework that handles a variety of languages for screenshot analysis. For the seamless discovery of tags, it's necessary to consider the knowledge around the extracted data as well. We particularly focused on the retrieval of screenshot images based on the image features, their textual contents, and related information.

This paper presents a complete workflow (from image pre-processing to tags retrieval) integrating existing open source methods and tools with our custom methods realized on-device. The on-device processing address the privacy concerns of the user. The background concepts are explained in Section ~\ref{sec:backg}. Section ~\ref{sec:process-pipeline} explains in detail the design of the overall pipeline while describing individual modules involved in detail. Section ~\ref{sec:tag_res} discuss the tag results. We end with future research directions in Section ~\ref{sec:conc}.

\section{Background} \label{sec:backg}
Tags can be generated based on image features and OCR. Image text is either machine printed like pdf or scene text like text present on objects. General steps for text extraction involve: Identify if any textual content exists, localize textual content and ultimately identify textual string from the identified area. For scanned documents, OCR has recognition rates > 99\%, however, for more complex or degraded images, processing techniques are still gathering research interest \cite{YeTextDet}, particularly in the context of natural scene images \cite{wang2012end}, where accuracy tends to drop around 60\%. Few traditional issues like skewed text and uneven illumination are mitigated in screenshots. However, it poses a different set of challenges like co-occurrence of icons and text and a vast set of fonts and layouts \cite{YeTextDet}. Further, screenshots represent a hybrid case study, mixing graphic and scene text in varying proportions over time, hence motivating the evaluation of existing techniques on a novel collection \cite{AgneseTextExt}.

The work presented in \cite{AgneseTextExt} shares some similarities with our current pipeline, however its more focused on extracting complete OCR, which is language dependent and server-based.

\section{Process Pipeline} \label{sec:process-pipeline}
Figure~\ref{fig:pipeline} shows the overview of the overall Process Pipeline. The various components are discussed in detail below.

\subsection{Dataset}
The procedures described in this paper are applied to a set of over 460 screenshots collected from 10 individuals. Ground truth tags are created for these screenshots by 2 trained human annotators. Screenshots being actual user data provide a meaningful evaluation of the solution developed. Apart from this, for training individual modules, various open source datasets in conjugation with our custom dataset are used, details of which will be provided in the respective modules.

\subsection{Image Pre-Processing}
A method is devised to process the raw screenshots so that the OCR engine could more easily distinguish textual content from the background. Details of which are as below:
\begin{itemize}
\item {\it {Filtering}}: Bilateral filtering has been applied in the image to reduce unwanted noise and smoothen the edges. This filter helps to keep the edge sharp. Before applying the filter, the alpha channel should be removed from the image.
\item {\it {Conversion to grayscale}}: Conversion of the images from RGB to grayscale is a prerequisite to binarization, which ultimately leads to better discrimination of the foreground from the background (i.e., the end goal of text/object detection) \cite{AgneseTextExt}.
\item {\it {Binarization}}: Binarization methods are deployed to convert grayscale images to binary format (i.e., black and white). We have written our custom method to process binarization on an image as shown in Figure~\ref{fig:binarization}.
\begin{figure}
  \centering
  \includegraphics[width=\linewidth]{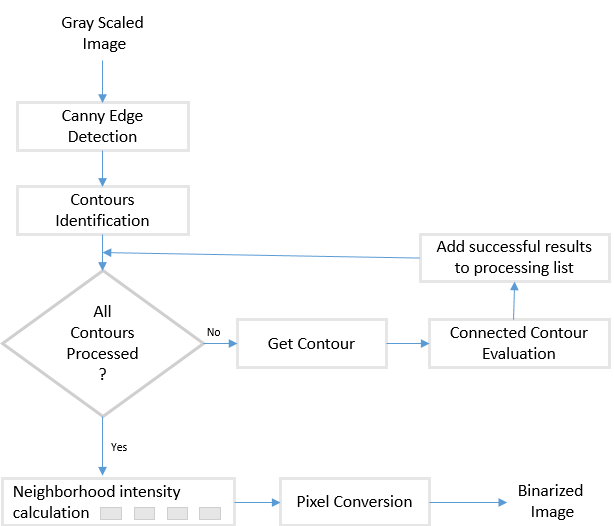}
  \caption{Binarization Flow Diagram}
  \label{fig:binarization}
\end{figure}
In this process, first edges are detected using canny edge on grayscale image. Further, these edges have been used to find out contours. With those contours, the connected components are evaluated and it is decided whether those points have to be processed further or not. For points that have to be processed, the intensity of all neighborhood pixels is calculated and accordingly decision is taken to convert that pixel into white or black. This task is time consuming. Hence, to optimize, processing is done in four different threads, which speeds up the process of image binarization.
\end{itemize}
Skew estimation step is avoided as screenshots have mostly horizontal content and skipping this step leads to an improvement in processing time.

\subsection{Script Detection}
\begin{figure}
  \centering
  \includegraphics[width=\linewidth]{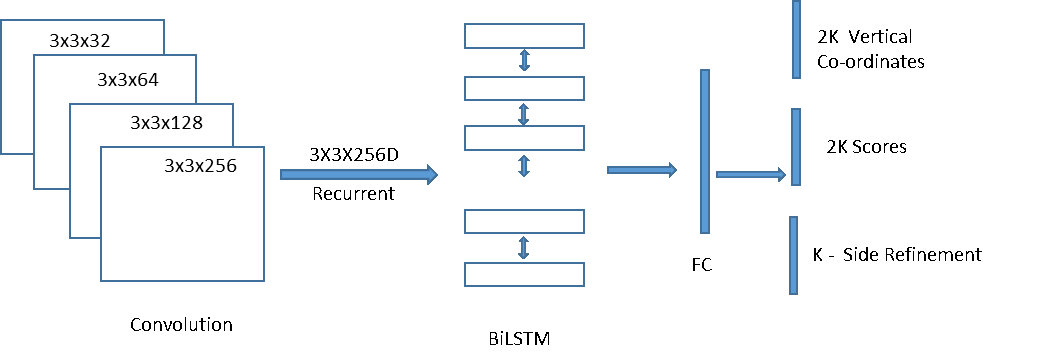}
  \caption{Architecture of the customized Connectionist Text Proposal Network (CTPN)}
  \label{fig:ctpn}
\end{figure}
It is important to detect the script used in the textual content for performing text extraction. Therefore, binarized image is furthered in the pipeline for Script Detection. This involves text localization and script identification.  Text localization detects bounding boxes of text regions. This is performed using CTPN \cite{tian2016ctpn}. It supports multilingual text localization. CTPN was trained on eleven different scripts. The modified network architecture of CTPN has been represented in Figure~\ref{fig:ctpn}. The network was modified to utilize a four layered convolution neural network for performing image feature extraction. The BiLSTM part was used as it was proposed in the Connectionist Text Proposal Network. The identified bounding boxes were further passed in the pipeline for the detection of the script. The problem of script detection is modeled as a classification problem and a four layer CNN based classifier is trained to classify the text image blocks as a certain script. To train the neural network a diverse dataset is created from ICDAR 2013 and ICDAR 2015 datasets, which provide the text box co-ordinates. The corresponding regions are cropped from the images and the dataset for training the CNN model is generated. This dataset is further augmented by synthetically generated data. On a set of 180 fonts, text contents of various fonts were rendered on a background. The background was generated using crop regions of certain natural images that were extracted from the ImageNet dataset. The text while rendering on the background image, is randomly skewed and blurred to generate a diverse dataset. A lightweight CNN with four layers was trained, thus obtaining an accuracy of 90\%. The model was trained to detect 11 scripts, namely Latin, Hangul, Chinese, Arabic, Devanagari, Bengali, Telugu, Tamil, Gujarati, Kannada and Malayalam. The metrics are summarized in Table~\ref{tab:script_met}.

\begin{table}[h]
  \caption{Metrics of Script Detection}
  \label{tab:script_met}
  \begin{tabular}{ccccc}
  \toprule
    &Model&Dataset&Size&Accuracy\\
  \midrule
    Text localization&CTPN&ICDAR&4MB&91\\
    Script Identification&CNN&Custom&5.3MB&90\\
  \bottomrule
  \end{tabular}
\end{table}

\subsection{Latin script OCR Engine}
After Script Detection, if Latin Script is detected, the pre-processed image is fed to a Latin script OCR Engine. MLKit\footnote{\url{https://developers.google.com/ml-kit/}} is used for the same. MLKit is a mobile SDK from google, which provides on-device text extraction for Latin. For MLKit, to accurately recognize text, input images must contain text that is represented by sufficient pixel data. Ideally, for Latin text, each character should be at least 16x16 pixels. The MLKit API needs to be provided with the image and a listener is set for the result.  If the text recognition operation succeeds, a FirebaseVisionText object will be passed to the listener.

A FirebaseVisionText object contains the full text recognized in the image and zero or more Text Block objects. Each Text Block represents a rectangular block of text, which contains zero or more Line objects. Each Line object contains zero or more Element objects, which represent words and word-like entities (dates, numbers, and so on). For each Text Block, Line, and Element object, the text recognized in the region and the bounding coordinates of the region is extracted. The MLKit on-device model size is approximately 10 MB.

\subsection{Non-Latin script OCR Engine}
After Script Detection, if non-Latin script is detected, the pre-processed image is fed to a Non-Latin script OCR Engine. A modified and on-device port of Tesseract\footnote{\url{https://github.com/tesseract-ocr/tesseract}} is used for the same. Tesseract latest version (v4.0) is used in our research as OCR performance and accuracy is better with this version. It uses LSTM based OCR Engine. The same pre-trained OCR Architecture is used as is and without any re-training.

Tesseract 4.0 is ported to android and native library code is written to use tesseract methods on Android devices. Apart from this, to improve the inference time of OCR, we did a major enhancement which has improved the inference time of OCR by 300 percent. Normal matrix dot product is replaced by neon optimized dot product.

We are passing binarized images with our pre-processing steps which achieves an accuracy of more than 93 percent on our custom data set. Char error rate and word error rate is less than 7 percent. Tesseract on-device port is approximately 3 MB. For each language supported, the data file is around 400-600 KB.

\subsection{Text to Language}
Text detected using the OCR engine is fed to the language detector module. Script determination is straightforward as different scripts fall under different Unicode ranges. However, identifying language poses a challenge as multiple languages in a script share the same set of characters (e.g. English, Spanish, and Italian in Latin script). Hence, Unicode cannot be a basis for disambiguating language given a script. Over the years, various methods have been proposed for Language Detection including statistical models \cite{mioulet2015language, lui2014automatic} and machine learning based models \cite{IvanaLangDet, kulmizev2017power}.

We evaluated existing solutions (MLKit, Apache OpenNLP, FastText) on LIGA Twitter Dataset for five Latin Languages (English, Spanish, Italian, French and German. However, these solutions did not satisfy the required performance and accuracy constraints.

Hence, we propose a Language Detector based on character co-occurrence statistics, which can identify language with 98.9\% accuracy with the library size of 500 KB. Table~\ref{tab:lang_det} summarizes accuracy of existing and proposed solutions.

\begin{table}[h]
  \caption{Language Detection-Accuracy}
  \label{tab:lang_det}
  \begin{tabular}{ccccc}
  \toprule
    &OpenNLP&MLKit&FastText&Proposed\\
  \midrule
    English&97.7&99.8&99.0&99.1\\
    Spanish&83.9&96.6&97.3&98.4\\
    Italian&93.5&96.2&98.1&98.5\\
    French&81.4&94.6&97.4&99.4\\
    German&94.9&96.1&99.4&98.6\\
  \midrule
    Overall&89.14&96.6&98.3&98.9\\
  \bottomrule
  \end{tabular}
\end{table}

Various studies\footnote{\url{http://practicalcryptography.com/cryptanalysis/letter-frequencies-various-languages/english-letter-frequencies}, \url{http://practicalcryptography.com/cryptanalysis/letter-frequencies-various-languages/spanish-letter-frequencies}} \cite{grigas2018letter, quaresma2008frequency} have been performed on language statistics, which involve analyzing the frequencies of Character n-gram (unigram, bigram and trigram) for various languages falling under the same script. It has been observed that most / least frequently occurring character n-grams are different for different languages and hence can be used to distinguish languages. A n-gram based Character Language Model (CLM) is built for each language to encode information about character n-gram frequencies into probabilities such that a value $\displaystyle P$ corresponding to n-gram ($\displaystyle a_{1} a_{2} \ldots a_{n}$) denoting the probability that n-gram belongs to the corresponding language is given by:
\begin{equation}
P( a_{n} | a_{1} \ a_{2} \dotsc a_{n-1}) \ =\ \frac{C( a_{1} \ a_{2} \dotsc a_{n} \ )}{C( a_{1} \ a_{2} \dotsc a_{( n-1)} \ )}
\end{equation}
where C is the count of the sequence of characters as observed from the corpus.

For building the CLM, data is crawled from various sources to have a complete representation of language. Smoothing techniques are applied to avoid zero frequency error. Finally, CLM is compressed and encrypted to ensure that memory constraints are satisfied. The size of compressed CLM is \textasciitilde30kB for each language. Further, it is observed that along with n-gram probability, the position of character n-gram within the word also plays an important role in the language identification task. N-gram probabilities are affected most when n-gram occurs as prefix (start of word) or suffix (end of word). Position specific probabilities are encoded in CLM by incorporating space (\_) as a valid character while building CLM. 

For word 'whether', learned character n-grams are:
\begin{verbatim}
  Trigrams: _wh, whe, het, eth, her, er_
  Bigrams: _w, wh, he, et, th, he, er, r_
\end{verbatim}

N-gram probability values stored in CLM are used to calculate the probability that given text belongs to a particular language $\displaystyle L_{i}$.
\begin{equation}
log( P( w_{1} \ w_{2} \dotsc w_{n} \ |\ L_{i})) =log( P( w_{1})) +log( P( w_{2})) +\cdots \ +log( P( w_{n}))
\end{equation}
\begin{equation}
P( w) =P( c_{1} \ c_{2} \ c_{3} \dotsc c_{n}) =P( c_{1}) P( c_{1} \ c_{2}) P( c_{1} \ c_{2} \ c_{3}) \dotsc P( c_{( n-2)} \ c_{( n-1)} \ c_{n})
\end{equation}
where $\displaystyle w_{1} w_{2} \ldots w_{n}$ is given text represented as word sequence. Each word is represented as a character sequence $\displaystyle c_{1} c_{2} \ldots c_{n}$ with space appended at start and end of word.

In this way, the probability of sentence belonging to each of the languages is calculated, $\displaystyle L_{i} \in \{L\}$ and the language with the highest probability value is selected.

Table~\ref{tab:langdetector} summarizes the result of language detector evaluation on the LIGA dataset in the form of the confusion matrix. Entry corresponding to [$\displaystyle L_{i}$, $\displaystyle L_{j}$] in the confusion matrix denotes the number of sentences that actually belong to language $\displaystyle L_{i}$ and have been classified as belonging to language $\displaystyle L_{j}$.

\begin{table}[h]
  \caption{Language Detection - Confusion Matrix}
  \label{tab:langdetector}
  \begin{tabular}{r*{4}{p{0.8cm}}p{0.9cm}l}
    \toprule
    & English & Spanish & Italian & French & German & Accuracy \\
    \midrule
    English & 1492 & 1 & 2 & 9 & 1 & 99.1 \\
    Spanish & 9 & 1537 & 5 & 9 & 2 & 98.4 \\
    Italian & 17 & 6 & 1516 & 0 & 0 & 98.5 \\
    French & 4 & 2 & 2 & 1542 & 1 & 99.4 \\
    German & 15 & 1 & 0 & 4 & 1459  & 98.6 \\
    \midrule
    Total & & & & & & 98.9 \\
    \bottomrule
  \end{tabular}
\end{table}

\subsection{Keyword Extraction}
\begin{table*}[htb]
  \caption{Metrics of English NER model (Bi-LSTM + CRF}
  \label{tab:engner}
  \begin{tabular}{lp{2.7cm}p{3.5cm}p{3.5cm}p{2cm}l}
    \toprule
    Model & Dataset & Word Embedding Dimension, Character Embedding Dimension & Num units in word LSTM, Num units in char LSTM & Quantized Model Size & F1 Score \\
    \midrule
    BiLSTM + CRF & CoNLL-2003 & 300, 100 & 300, 100 & 8.6 MB & 90.4  \\
    BiLSTM + CRF & CoNLL-2003 + External Dataset & 300, 100 & 300, 100 & 8.6 MB & 91.2  \\
    BiLSTM + CRF & CoNLL-2003 + External Dataset & 100, 50 & 100, 50 & 2.7 MB & 90.0  \\
    \bottomrule
  \end{tabular}
\end{table*}
\textbf{English/Spanish:} For keyword extraction, we use Named Entity Recognition (NER) which is a subtask of information extraction that seeks to locate and classify named entities in text into pre-defined categories such as the names of person, organizations, locations, etc. What makes this problem non-trivial is that many entities, like names or organizations are just made-up names for which any prior knowledge is not known. Thus, a deep learning solution is required that will extract contextual information from the sentence, just as if humans do.

For building NER for English and Spanish, a model similar to Lample et al. \cite{LampleNER} and Ma and Hovy \cite{MaHovy} is used. Firstly, a Bi-LSTM is trained to get character embeddings from the training dataset, to get a character-based representation of each word. Next, this is concatenated with standard GloVe\footnote{\url{https://nlp.stanford.edu/projects/glove}} (300 dimension vectors trained on a 6 billion corpus of Wikipedia 2014 and Gigaword5) word vector representation. This gives us the contextual representation of each word. Then, a Bi-LSTM is run on each sentence represented by the above contextual representation. This final output from the model is decoded with a linear chain CRF using Viterbi algorithm (Tensorflow's crf.viterbi\_decode method)\footnote{\url{https://www.tensorflow.org/api_docs/python/tf/contrib/crf/viterbi_decode}}. For on-device inference, the same Viterbi decode function is implemented in Java to be run on android devices and get the final output. The same model is used for Spanish and other Latin languages. The model is quantized to reduce its size and put it on-device without increasing the app size much.

For English keyword detection from NER, the tagged dataset from the CoNLL-2003 shared task is combined with the dataset collected in-house from 20 users and tagged manually. For Spanish, the CoNLL-2002 shared task dataset for NER is used. The various parameters and metrics of English NER are listed in Table~\ref{tab:engner}. The final on-device English NER model had a F1 score 90, while the Spanish NER model had a F1 score of 78.

\textbf{Hindi:} For keyword extraction, a java implementation of Single Classification Ripple Down Rules (SCRDR) tree of transformation rules for part-of-speech (PoS) tagging task from Nyugen et al. \cite{nguyen2014rdrpostagger} is used. SCRDR PoS Tagger obtains very fast tagging speed, achieves a competitive accuracy in comparison to the state-of-the-art results, and hence is the perfect choice for on-device PoS tagging. It also supports pre-trained models for fine-grained PoS tagging. For our purposes Rules and Dictionary dataset for Hindi Language from Universal Dependencies (UD) v2.0\footnote{\url{https://universaldependencies.org}} is used which has a PoS accuracy of 94.91\% on Hindi PoS tagging. The rules and dictionary files are approximately 500 KB in size.

\subsection{Scene Tag Classification}
Image Classification problem is the task of assigning an input image one label from a fixed set of categories to an input image. A given image is classified into a few of the 10000 output labels.
The input image is fed to MobileNet \cite{AndrewMobileNets}, for image classification. MobileNet is chosen for this purpose because of its small size and faster speed, which helps in on-device deployment and inference. A pre-trained MobileNet tensorflow model\footnote{\url{http://download.tensorflow.org/models/mobilenet_v1_2018_08_02/mobilenet_v1_1.0_224.tgz}} trained on ILSVRC-2012-CLS image classification dataset\footnote{\url{http://www.image-net.org/challenges/LSVRC/2012}} is used. The model has a top 1 accuracy of 75\% and top 5 accuracy of 92.5\%. The model has more than 10000 output classes.

\subsection{Related Keywords}
In this section, how the extracted tags from the content are expanded to include more related tags, is discussed. This facilitates the user to search for content with semantically similar or related search terms. To achieve this, the utilities of a Knowledge Graph to find connected entities for a given entity (tag in our case) are exploited. Exploring the connections and their types, the tag set can be expanded to a larger and more relevant set. A neural model is employed to find related entities of a given entity in the Knowledge Graph. 

The model takes as input an entity in the form of a tag. The Knowledge Graph stores real world information in a directed multi-relational structured graph consisting of entities and directed edges. Each unit of knowledge is represented as a triplet \{head, relation, tail\} where the head and tail are the connected entities while the edge connecting them corresponds to the relationship between them. For example, a piece of information like "Tokyo is the capital of Japan" in triplet representation is \{Tokyo, \textit{isCapitalOf}, Japan\}. For simplification, the nature of relations is not considered to be a factor in determining connected tags. Therefore, all the relations in the triplets is replaced with a common \textit{related\_to} relation, while preserving the directed property of the graph.

A CNN is developed on the design proposed by \cite{ZhaoGraph}. The network learns the embeddings of entities $\displaystyle e\in E$, relations $\displaystyle l\in L$ and score functions together, similar to conventional translation-based models like TransE. Based on a negative sampling method, for a given positive training set S, a negative training Set S' is prepared by randomly replacing head or tail (but not both at the same time) as shown in equation~\ref{eq:rel1}:
\begin{equation} \label{eq:rel1}
S'( h,l,t) =\{( h',l,t) |h'\in E\} \cup \{( h,l,t') \ |\ t'\in E\}
\end{equation}

\begin{figure}[H]
  \centering
  \includegraphics[width=\linewidth]{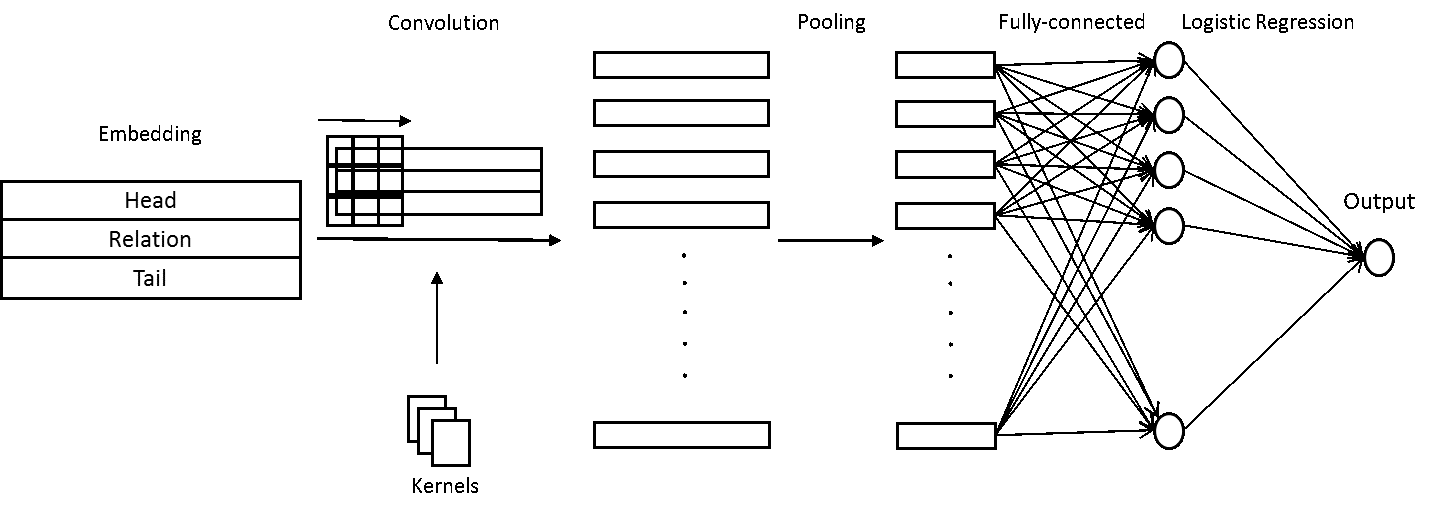}
  \caption{CNN Related Keywords Model}
  \label{fig:cnnrelated}
\end{figure}

The score function is trained to evaluate the positive triplets with a higher score and the negative triplets with low scores. The CNN layers as illustrated in Figure~\ref{fig:cnnrelated} include the (i) embedding layer (ii) one convolution layer (iii) max pooling layer (iv) a fully connected layer and a (v) logistic regression layer.

The on-device model size is approximately 2.6 MB. The mean rank of the model is 68 and hits@10 is 94.5\%. In Table~\ref{tab:comp_cnn}, we have compared the performance of the network with other contemporary approaches on two standard datasets, Freebase \cite{bollacker2008freebase} and WordNet \cite{miller1995wordnet} on the standard link prediction problem.
\begin{table}
  \caption{Comparison of our CNN model with state of the art methods on link prediction}
  \label{tab:comp_cnn}
  \begin{tabular}{lp{0.6cm}c|lp{0.6cm}c}
    \toprule
    FB15k&Mean Rank&Hits@10&WN18&Mean Rank&Hits@10\\
    \midrule
    TransE&117&45.3&TransE&266&86.7\\
    TransH&90&66.8&TransH&298&84.9\\
    TransR&87&67.5&TransR&278&85.6\\
    PTransE&55&79.5&PTransE&244&89.9\\
    ProjE&39&86.7&ProjE&198&94.2\\
    CNN&61&93.4&CNN&54&95.6\\
    \bottomrule
  \end{tabular}
\end{table}
\begin{figure*}
  \centering
  \includegraphics[width=\linewidth]{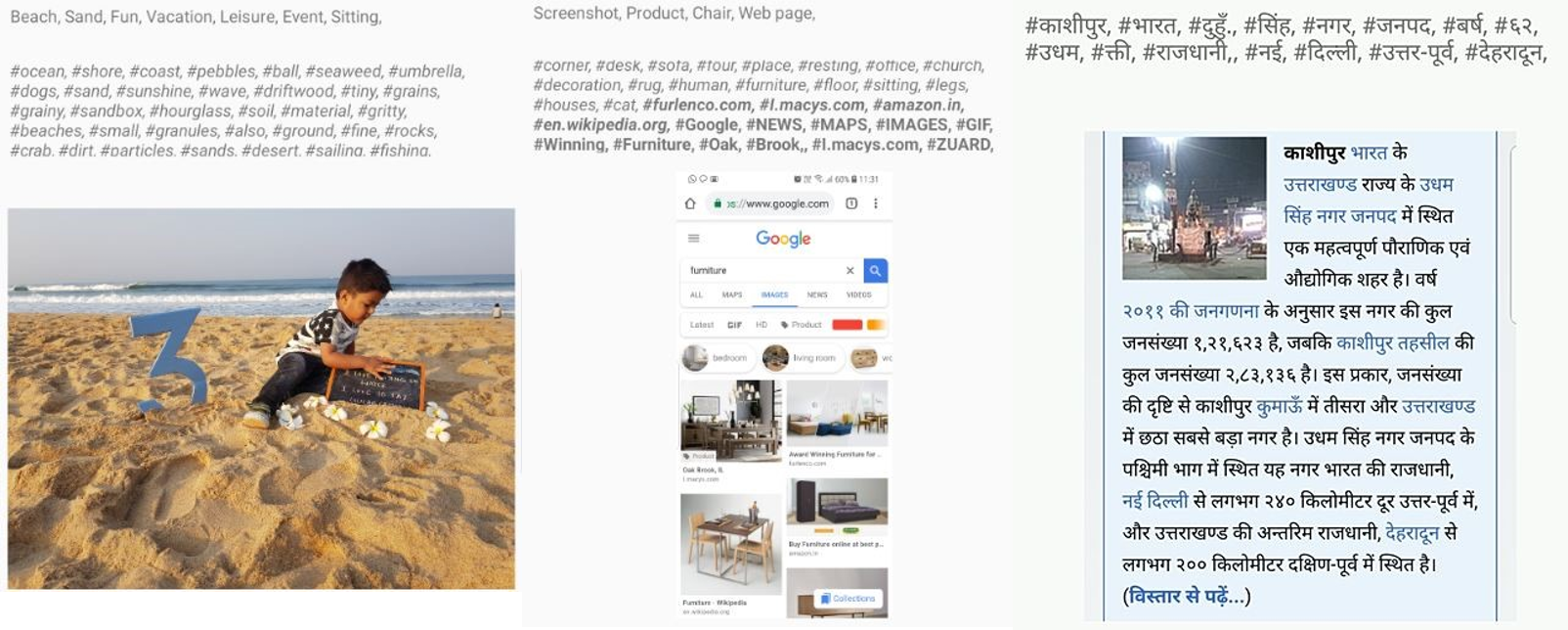}
  \caption{Tags generated on few sample images}
  \label{fig:results}
\end{figure*}

\subsection{Ranker}
After the above process, multiple tags related to the image are generated which need to be ranked before displaying to end user. Tag score for individual tags are calculated as per below equation,
\begin{equation}
\begin{array}{l}
Tag\ score=prob( scene\ tag) +0.8\ \ast prob( ocr\ tag) +\\
\left( prob( parent\ tag) \ \ast \ e^{-prob( related\ tag)} \ \forall \ related\ tags\right)
\end{array}
\end{equation}
where parent tag is the scene or OCR tag, from which the related tag is derived from.

\section{Tag Results and Discussion} \label{sec:tag_res}
\begin{figure}
  \centering
  \includegraphics[width=\linewidth]{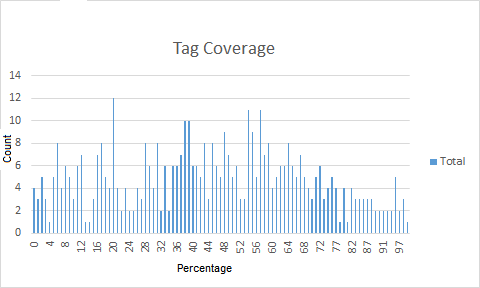}
  \caption{Experimental results on test dataset of 460 images}
  \label{fig:tag_results}
\end{figure}
Figure ~\ref{fig:results} shows tags generated on a few sample images. We compared tags generated from our solution with ground truth tags using Tag coverage which is the percentage of tags covered by the proposed method against ground truth tags as shown in Figure~\ref{fig:tag_results}. We verified that 63\% test cases partially passed and approximately 12\% passed as shown in Table~\ref{tab:consresult}.

\begin{table}
  \caption{Consolidated Result}
  \label{tab:consresult}
  \begin{tabular}{ccc}
    \toprule
    & Tag Coverage & Count \\
    \midrule
    Pass & \textgreater 75\% & 55 \\
    Partially Pass & \textgreater 25\% & 290 \\
    Fail & \textless 25\% & 115 \\
     \bottomrule
  \end{tabular}
\end{table}

\section{Conclusions and Future Work} \label{sec:conc}
This paper introduced a complete pipeline for tags extraction and retrieval from smartphone screenshots. The pipeline is based on in-house developed modules as described in Section \ref{sec:process-pipeline} apart from MLKit and Tesseract OCR. The accuracy of tags is evaluated and shown how tags can be generated automatically based on OCR and image analysis. Detailed analysis of results suggests that further improvements are possible in the pipeline both by image pre-processing and individual modules accuracy improvements.

Future work may include region of interest identification, which should increase the solution's robustness in the presence of mixed graphic contents. Pre-processing may be particularly important for text extraction from multi-window screenshots. Thus methods that effectively and accurately partition and segment the screen will need to be developed and refined. There is a major area of improvement for related tags generation in which case, dataset needs to be expanded to get the result which is more relevant. These additions will expand computation complexity and time and therefore need to be evaluated with respect to benefit. Furthermore, this pipeline can be expanded to suggest tags based on video input.

To summarize, this paper describes the procedures followed for extracting and generating tags from mobile screenshots on-device, which also address privacy concerns of users of having their data being sent to the server for tag suggestions. These initial results are promising as the user can search for a screenshot with the help of the tags, thus contributing to easing user life.

%
\bibliographystyle{ACM-Reference-Format}
\bibliography{references}


\begin{thebibliography}{19}


\ifx \showCODEN    \undefined \def \showCODEN     #1{\unskip}     \fi
\ifx \showDOI      \undefined \def \showDOI       #1{#1}\fi
\ifx \showISBNx    \undefined \def \showISBNx     #1{\unskip}     \fi
\ifx \showISBNxiii \undefined \def \showISBNxiii  #1{\unskip}     \fi
\ifx \showISSN     \undefined \def \showISSN      #1{\unskip}     \fi
\ifx \showLCCN     \undefined \def \showLCCN      #1{\unskip}     \fi
\ifx \shownote     \undefined \def \shownote      #1{#1}          \fi
\ifx \showarticletitle \undefined \def \showarticletitle #1{#1}   \fi
\ifx \showURL      \undefined \def \showURL       {\relax}        \fi
\providecommand\bibfield[2]{#2}
\providecommand\bibinfo[2]{#2}
\providecommand\natexlab[1]{#1}
\providecommand\showeprint[2][]{arXiv:#2}

\bibitem[\protect\citeauthoryear{Balazevic, Braun, and M{\"{u}}ller}{Balazevic
  et~al\mbox{.}}{2016}]%
        {IvanaLangDet}
\bibfield{author}{\bibinfo{person}{Ivana Balazevic},
  \bibinfo{person}{Mikio~Ludwig Braun}, {and} \bibinfo{person}{Klaus{-}Robert
  M{\"{u}}ller}.} \bibinfo{year}{2016}\natexlab{}.
\newblock \showarticletitle{Language Detection For Short Text Messages In
  Social Media}.
\newblock \bibinfo{journal}{\emph{CoRR}}  \bibinfo{volume}{abs/1608.08515}
  (\bibinfo{year}{2016}).
\newblock
\showeprint[arxiv]{1608.08515}
\urldef\tempurl%
\url{http://arxiv.org/abs/1608.08515}
\showURL{%
\tempurl}


\bibitem[\protect\citeauthoryear{Bollacker, Evans, Paritosh, Sturge, and
  Taylor}{Bollacker et~al\mbox{.}}{2008}]%
        {bollacker2008freebase}
\bibfield{author}{\bibinfo{person}{Kurt Bollacker}, \bibinfo{person}{Colin
  Evans}, \bibinfo{person}{Praveen Paritosh}, \bibinfo{person}{Tim Sturge},
  {and} \bibinfo{person}{Jamie Taylor}.} \bibinfo{year}{2008}\natexlab{}.
\newblock \showarticletitle{Freebase: a collaboratively created graph database
  for structuring human knowledge}. In \bibinfo{booktitle}{\emph{Proceedings of
  the 2008 ACM SIGMOD international conference on Management of data}}. AcM,
  \bibinfo{pages}{1247--1250}.
\newblock


\bibitem[\protect\citeauthoryear{Chiatti, Cho, Gagneja, Yang, Brinberg,
  Roehrick, Choudhury, Ram, Reeves, and Giles}{Chiatti et~al\mbox{.}}{2018}]%
        {AgneseTextExt}
\bibfield{author}{\bibinfo{person}{Agnese Chiatti}, \bibinfo{person}{Mu~Jung
  Cho}, \bibinfo{person}{Anupriya Gagneja}, \bibinfo{person}{Xiao Yang},
  \bibinfo{person}{Miriam Brinberg}, \bibinfo{person}{Katie Roehrick},
  \bibinfo{person}{Sagnik~Ray Choudhury}, \bibinfo{person}{Nilam Ram},
  \bibinfo{person}{Byron Reeves}, {and} \bibinfo{person}{C~Lee Giles}.}
  \bibinfo{year}{2018}\natexlab{}.
\newblock \showarticletitle{Text extraction and retrieval from smartphone
  screenshots: building a repository for life in media}.
\newblock \bibinfo{journal}{\emph{CoRR}}  \bibinfo{volume}{abs/1801.01316}
  (\bibinfo{year}{2018}), \bibinfo{pages}{948--955}.
\newblock


\bibitem[\protect\citeauthoryear{Chiatti, Yang, Brinberg, Jung~Cho, Gagneja,
  Ram, Reeves, and Lee~Giles}{Chiatti et~al\mbox{.}}{2017}]%
        {ChiattiTextExt}
\bibfield{author}{\bibinfo{person}{Agnese Chiatti}, \bibinfo{person}{Xiao
  Yang}, \bibinfo{person}{Miriam Brinberg}, \bibinfo{person}{Mu Jung~Cho},
  \bibinfo{person}{Anupriya Gagneja}, \bibinfo{person}{Nilam Ram},
  \bibinfo{person}{Byron Reeves}, {and} \bibinfo{person}{C Lee~Giles}.}
  \bibinfo{year}{2017}\natexlab{}.
\newblock \showarticletitle{Text Extraction from Smartphone Screenshots to
  Archive in situ Media Behavior}. \bibinfo{pages}{1--4}.
\newblock
\urldef\tempurl%
\url{https://doi.org/10.1145/3148011.3154468}
\showDOI{\tempurl}


\bibitem[\protect\citeauthoryear{Grigas and
  Ju{\v{s}}kevi{\v{c}}ien{\.e}}{Grigas and
  Ju{\v{s}}kevi{\v{c}}ien{\.e}}{2018}]%
        {grigas2018letter}
\bibfield{author}{\bibinfo{person}{Gintautas Grigas} {and}
  \bibinfo{person}{Anita Ju{\v{s}}kevi{\v{c}}ien{\.e}}.}
  \bibinfo{year}{2018}\natexlab{}.
\newblock \showarticletitle{Letter Frequency Analysis of Languages Using Latin
  Alphabet}.
\newblock \bibinfo{journal}{\emph{International Linguistics Research}}
  \bibinfo{volume}{1}, \bibinfo{number}{1} (\bibinfo{year}{2018}),
  \bibinfo{pages}{p18--p18}.
\newblock


\bibitem[\protect\citeauthoryear{Howard, Zhu, Chen, Kalenichenko, Wang, Weyand,
  Andreetto, and Adam}{Howard et~al\mbox{.}}{2017}]%
        {AndrewMobileNets}
\bibfield{author}{\bibinfo{person}{Andrew~G. Howard}, \bibinfo{person}{Menglong
  Zhu}, \bibinfo{person}{Bo Chen}, \bibinfo{person}{Dmitry Kalenichenko},
  \bibinfo{person}{Weijun Wang}, \bibinfo{person}{Tobias Weyand},
  \bibinfo{person}{Marco Andreetto}, {and} \bibinfo{person}{Hartwig Adam}.}
  \bibinfo{year}{2017}\natexlab{}.
\newblock \showarticletitle{MobileNets: Efficient Convolutional Neural Networks
  for Mobile Vision Applications}.
\newblock \bibinfo{journal}{\emph{CoRR}}  \bibinfo{volume}{abs/1704.04861}
  (\bibinfo{year}{2017}).
\newblock
\showeprint[arxiv]{1704.04861}
\urldef\tempurl%
\url{http://arxiv.org/abs/1704.04861}
\showURL{%
\tempurl}


\bibitem[\protect\citeauthoryear{Kulmizev, Blankers, Bjerva, Nissim, van Noord,
  Plank, and Wieling}{Kulmizev et~al\mbox{.}}{2017}]%
        {kulmizev2017power}
\bibfield{author}{\bibinfo{person}{Artur Kulmizev}, \bibinfo{person}{Bo
  Blankers}, \bibinfo{person}{Johannes Bjerva}, \bibinfo{person}{Malvina
  Nissim}, \bibinfo{person}{Gertjan van Noord}, \bibinfo{person}{Barbara
  Plank}, {and} \bibinfo{person}{Martijn Wieling}.}
  \bibinfo{year}{2017}\natexlab{}.
\newblock \showarticletitle{The power of character n-grams in native language
  identification}. In \bibinfo{booktitle}{\emph{Proceedings of the 12th
  Workshop on Innovative Use of NLP for Building Educational Applications}}.
  \bibinfo{pages}{382--389}.
\newblock


\bibitem[\protect\citeauthoryear{Lample, Ballesteros, Subramanian, Kawakami,
  and Dyer}{Lample et~al\mbox{.}}{2016}]%
        {LampleNER}
\bibfield{author}{\bibinfo{person}{Guillaume Lample}, \bibinfo{person}{Miguel
  Ballesteros}, \bibinfo{person}{Sandeep Subramanian}, \bibinfo{person}{Kazuya
  Kawakami}, {and} \bibinfo{person}{Chris Dyer}.}
  \bibinfo{year}{2016}\natexlab{}.
\newblock \showarticletitle{Neural Architectures for Named Entity Recognition}.
\newblock \bibinfo{journal}{\emph{CoRR}}  \bibinfo{volume}{abs/1603.01360}
  (\bibinfo{year}{2016}).
\newblock
\showeprint[arxiv]{1603.01360}
\urldef\tempurl%
\url{http://arxiv.org/abs/1603.01360}
\showURL{%
\tempurl}


\bibitem[\protect\citeauthoryear{Lui, Lau, and Baldwin}{Lui
  et~al\mbox{.}}{2014}]%
        {lui2014automatic}
\bibfield{author}{\bibinfo{person}{Marco Lui}, \bibinfo{person}{Jey~Han Lau},
  {and} \bibinfo{person}{Timothy Baldwin}.} \bibinfo{year}{2014}\natexlab{}.
\newblock \showarticletitle{Automatic detection and language identification of
  multilingual documents}.
\newblock \bibinfo{journal}{\emph{Transactions of the Association for
  Computational Linguistics}}  \bibinfo{volume}{2} (\bibinfo{year}{2014}),
  \bibinfo{pages}{27--40}.
\newblock


\bibitem[\protect\citeauthoryear{Ma and Hovy}{Ma and Hovy}{2016}]%
        {MaHovy}
\bibfield{author}{\bibinfo{person}{Xuezhe Ma} {and} \bibinfo{person}{Eduard~H.
  Hovy}.} \bibinfo{year}{2016}\natexlab{}.
\newblock \showarticletitle{End-to-end Sequence Labeling via Bi-directional
  LSTM-CNNs-CRF}.
\newblock \bibinfo{journal}{\emph{CoRR}}  \bibinfo{volume}{abs/1603.01354}
  (\bibinfo{year}{2016}).
\newblock
\showeprint[arxiv]{1603.01354}
\urldef\tempurl%
\url{http://arxiv.org/abs/1603.01354}
\showURL{%
\tempurl}


\bibitem[\protect\citeauthoryear{Miller}{Miller}{1995}]%
        {miller1995wordnet}
\bibfield{author}{\bibinfo{person}{George~A Miller}.}
  \bibinfo{year}{1995}\natexlab{}.
\newblock \showarticletitle{WordNet: a lexical database for English}.
\newblock \bibinfo{journal}{\emph{Commun. ACM}} \bibinfo{volume}{38},
  \bibinfo{number}{11} (\bibinfo{year}{1995}), \bibinfo{pages}{39--41}.
\newblock


\bibitem[\protect\citeauthoryear{Mioulet, Garain, Chatelain, Barlas, and
  Paquet}{Mioulet et~al\mbox{.}}{2015}]%
        {mioulet2015language}
\bibfield{author}{\bibinfo{person}{Luc Mioulet}, \bibinfo{person}{Utpal
  Garain}, \bibinfo{person}{Cl{\'e}ment Chatelain}, \bibinfo{person}{Philippine
  Barlas}, {and} \bibinfo{person}{Thierry Paquet}.}
  \bibinfo{year}{2015}\natexlab{}.
\newblock \showarticletitle{Language identification from handwritten
  documents}. In \bibinfo{booktitle}{\emph{2015 13th International Conference
  on Document Analysis and Recognition (ICDAR)}}. IEEE,
  \bibinfo{pages}{676--680}.
\newblock


\bibitem[\protect\citeauthoryear{Nguyen, Nguyen, Pham, and Pham}{Nguyen
  et~al\mbox{.}}{2014}]%
        {nguyen2014rdrpostagger}
\bibfield{author}{\bibinfo{person}{Dat~Quoc Nguyen}, \bibinfo{person}{Dai~Quoc
  Nguyen}, \bibinfo{person}{Dang~Duc Pham}, {and} \bibinfo{person}{Son~Bao
  Pham}.} \bibinfo{year}{2014}\natexlab{}.
\newblock \showarticletitle{RDRPOSTagger: A ripple down rules-based
  part-of-speech tagger}. In \bibinfo{booktitle}{\emph{Proceedings of the
  Demonstrations at the 14th Conference of the European Chapter of the
  Association for Computational Linguistics}}. \bibinfo{pages}{17--20}.
\newblock


\bibitem[\protect\citeauthoryear{Quaresma}{Quaresma}{2008}]%
        {quaresma2008frequency}
\bibfield{author}{\bibinfo{person}{Pedro Quaresma}.}
  \bibinfo{year}{2008}\natexlab{}.
\newblock \showarticletitle{Frequency analysis of the Portuguese language}.
\newblock \bibinfo{journal}{\emph{University of Coimbra, Portugal}}
  (\bibinfo{year}{2008}).
\newblock


\bibitem[\protect\citeauthoryear{Reeves, Ram, N.~Robinson, Cummings, Lee~Giles,
  Pan, Chiatti, Cho, Roehrick, Yang, Gagneja, Brinberg, Muise, Lu, Luo,
  Fitzgerald, and Yeykelis}{Reeves et~al\mbox{.}}{2019}]%
        {ReevesScreenomics}
\bibfield{author}{\bibinfo{person}{Byron Reeves}, \bibinfo{person}{Nilam Ram},
  \bibinfo{person}{Thomas N.~Robinson}, \bibinfo{person}{James Cummings},
  \bibinfo{person}{C Lee~Giles}, \bibinfo{person}{Jennifer Pan},
  \bibinfo{person}{Agnese Chiatti}, \bibinfo{person}{Mj Cho},
  \bibinfo{person}{Katie Roehrick}, \bibinfo{person}{Xiao Yang},
  \bibinfo{person}{Anupriya Gagneja}, \bibinfo{person}{Miriam Brinberg},
  \bibinfo{person}{Daniel Muise}, \bibinfo{person}{Yingdan Lu},
  \bibinfo{person}{Mufan Luo}, \bibinfo{person}{Andrew Fitzgerald}, {and}
  \bibinfo{person}{Leo Yeykelis}.} \bibinfo{year}{2019}\natexlab{}.
\newblock \showarticletitle{Screenomics : A Framework to Capture and Analyze
  Personal Life Experiences and the Ways that Technology Shapes Them}.
\newblock \bibinfo{journal}{\emph{Human-Computer Interaction}}
  (\bibinfo{date}{03} \bibinfo{year}{2019}), \bibinfo{pages}{1--52}.
\newblock
\urldef\tempurl%
\url{https://doi.org/10.1080/07370024.2019.1578652}
\showDOI{\tempurl}


\bibitem[\protect\citeauthoryear{Tian, Huang, He, He, and Qiao}{Tian
  et~al\mbox{.}}{2016}]%
        {tian2016ctpn}
\bibfield{author}{\bibinfo{person}{Zhi Tian}, \bibinfo{person}{Weilin Huang},
  \bibinfo{person}{Tong He}, \bibinfo{person}{Pan He}, {and}
  \bibinfo{person}{Yu Qiao}.} \bibinfo{year}{2016}\natexlab{}.
\newblock \showarticletitle{Detecting text in natural image with connectionist
  text proposal network}. In \bibinfo{booktitle}{\emph{European conference on
  computer vision}}. Springer, \bibinfo{pages}{56--72}.
\newblock


\bibitem[\protect\citeauthoryear{Wang, Wu, Coates, and Ng}{Wang
  et~al\mbox{.}}{2012}]%
        {wang2012end}
\bibfield{author}{\bibinfo{person}{Tao Wang}, \bibinfo{person}{David~J Wu},
  \bibinfo{person}{Adam Coates}, {and} \bibinfo{person}{Andrew~Y Ng}.}
  \bibinfo{year}{2012}\natexlab{}.
\newblock \showarticletitle{End-to-end text recognition with convolutional
  neural networks}. In \bibinfo{booktitle}{\emph{Proceedings of the 21st
  International Conference on Pattern Recognition (ICPR2012)}}. IEEE,
  \bibinfo{pages}{3304--3308}.
\newblock
\showISSN{1051-4651}


\bibitem[\protect\citeauthoryear{Ye and S.~Doermann}{Ye and
  S.~Doermann}{2015}]%
        {YeTextDet}
\bibfield{author}{\bibinfo{person}{Qixiang Ye} {and} \bibinfo{person}{David
  S.~Doermann}.} \bibinfo{year}{2015}\natexlab{}.
\newblock \showarticletitle{Text Detection and Recognition in Imagery: A
  Survey}.
\newblock \bibinfo{journal}{\emph{IEEE Transactions on Pattern Analysis and
  Machine Intelligence}}  \bibinfo{volume}{37} (\bibinfo{date}{06}
  \bibinfo{year}{2015}).
\newblock
\urldef\tempurl%
\url{https://doi.org/10.1109/TPAMI.2014.2366765}
\showDOI{\tempurl}


\bibitem[\protect\citeauthoryear{Zhao, Min, Shen, and Chakraborty}{Zhao
  et~al\mbox{.}}{2017}]%
        {ZhaoGraph}
\bibfield{author}{\bibinfo{person}{Feipeng Zhao},
  \bibinfo{person}{Martin~Renqiang Min}, \bibinfo{person}{Chen Shen}, {and}
  \bibinfo{person}{Amit Chakraborty}.} \bibinfo{year}{2017}\natexlab{}.
\newblock \showarticletitle{Convolutional Neural Knowledge Graph Learning}.
\newblock \bibinfo{journal}{\emph{CoRR}}  \bibinfo{volume}{abs/1710.08502}
  (\bibinfo{year}{2017}).
\newblock
\showeprint[arxiv]{1710.08502}
\urldef\tempurl%
\url{http://arxiv.org/abs/1710.08502}
\showURL{%
\tempurl}
\newblock
\shownote{Withdrawn.}


\end{thebibliography}

\end{document}